\documentclass{article} % For LaTeX2e
\usepackage{iclr2017_conference_customized,times}
\usepackage{natbib}
\usepackage{hyperref}
\usepackage{url}

\usepackage{amsfonts, amsmath}
\usepackage{graphics,epsfig}
\usepackage{caption}
\usepackage{subcaption}
\usepackage{xcolor,colortbl}
\title{Deep Learning of Robotic Tasks without a Simulator \\
	 using Strong and Weak Human Supervision}

\date{}

\author{Bar Hilleli \& Ran El-Yaniv \\
Department of Computer Science \\
Technion - Israel Institute of Technology \\
Haifa, Israel \\
\texttt{barh@campus.technion.ac.il} \\
\texttt{rani@cs.technion.ac.il} 
}

\begin{document}

\maketitle
\begin{abstract}
We propose a scheme for training a computerized agent to perform complex human tasks such as highway steering.
The scheme is designed to follow a natural learning process whereby a human instructor teaches a computerized trainee.
The learning process consists of five elements: (i) unsupervised feature learning; (ii) supervised imitation learning; (iii) supervised reward induction; (iv) supervised safety module construction; and (v) reinforcement learning.
We implemented the last four elements of the scheme using deep convolutional networks and applied it to successfully create a computerized agent capable of autonomous highway steering over the well-known racing game \emph{Assetto Corsa}.
We demonstrate that the use of the last four elements is essential to effectively carry out the steering task using vision alone, without access to a driving simulator internals, and operating in wall-clock time. 
This is made possible also through the introduction of a \emph{safety network}, a novel way for preventing the agent from performing catastrophic mistakes during the reinforcement learning stage.		
	
%We propose a scheme for training a computerized agent to perform complex human tasks such as highway steering.
%The scheme resembles natural teaching-learning processes used by humans to teach themselves and each other complex tasks, and consists of the following four stages.
%In the first stage the agent learns by itself an informative low-dimensional representations of raw input signals in an unsupervised learning manner.
%In the second stage the agent learns to mimic the human instructor using supervised learning so as to reach a basic performance level; the third stage is devoted to learning an instantaneous reward model. 
%Here, the (human) instructor observes (possibly in real time) the agent performing the task and provides reward feedback.
%During this stage the agent monitors both itself and the instructor feedback and learns a reward model using supervised learning. 
%This stage terminates when the reward model is sufficiently accurate.
%In the last stage a reinforcement learning algorithm is deployed to optimize the agent policy.
%The guidance reward signal in the reinforcement learning algorithm relies on the previously learned reward model.
%In addition, we introduce the idea of a \emph{safety network}, a novel way for preventing the agent from performing catastrophic mistakes during the RL stage.
%As a proof of concept for the proposed scheme, we designed a system consisting of deep convolutional neural networks, and applied it to successfully learn a computerized agent capable of autonomous highway steering over the well-known racing game \emph{Assetto Corsa}.
\end{abstract}

\section{Introduction}
% general motivation
Consider the task of designing a robot capable of performing a complex human task such as dishwashing, driving or clothes ironing.
Although these tasks are natural for adult humans, designing a hard-coded algorithm for such a robot can be a daunting challenge.
Difficulties in accurately modeling the robot and its interaction with the environment, creating  hand-crafted features from the high-dimensional sensor data, and the requirement that the robot be able to adapt to new situations are just a few of these obstacles.
In this paper we propose a general scheme that combines several learning techniques that might be used to tackle such challenges. 
As a proof a concept, we implemented the scheme's stages (currently without the initial unsupervised learning stage) and applied it to the challenging problem of autonomous highway steering.

%  IL and RL
Two of the most common approaches for robot learning are \emph{Imitation Learning} (IL) and \emph{Reinforcement Learning} (RL).
In imitation learning, which is also known as `behavioral cloning' or `learning from demonstrations', a human demonstrator performs the desired task with the goal of teaching a (robotic) agent to mimic her actions \citep{argall2009survey}.
The demonstrations are used to learn a mapping from a given world state, $s$, received via sensors, to a desired action, $a$, consisting of instructions to the agent's controllers.
Throughout this paper $s$ will be referred also as the "raw signal."
The objective in IL is to minimize the risk, in the supervised learning (SL) sense.
In RL the goal is to enable the agent to find its own policy, one that maximizes a value function defined in terms of certain guidance reward signals received during its interaction with the environment. 
IL and RL can be combined, as was recently proposed by \cite{taylor2011using}.
The idea is to start the reinforcement learning with an initial policy learned during a preceding IL stage.
This combined approach can significantly accelerate the RL learning process and minimize costly agent-environment interactions.   
In addition, deploying the RL algorithm after the IL stage compensates for the noisy demonstrations and extends the imitation strategy to previously unseen areas of the state space. 

% discribe the disadventages of the two methods and how combining them is very helpful
Noisy demonstrations in the IL stage typically result from inconsistent strategies adopted by the human demonstrator.
Consequently, the performance of an IL agent is limited by the quality of the observed demonstration.
RL has it drawbacks as well: it either requires a realistic simulation of the agent's interaction with the environment, or it requires operating the agent in a real-world environment, which can be quite costly.
Moreover, the sample complexity of RL can be large and the modeling of an effective reward function is often quite challenging, requiring expert insight and domain knowledge. 
The current state of affairs in robotic design using any technique leaves much to be desired, where ultimate goals such as domestic robot housekeepers remain futuristic.

% several modalities
% MODIFICATION: added paragraph
Consider the task of constructing a robot (or any computerized agent) whose goal is to perform a certain task based on raw signals obtained via sensors.
The sensors can be of several types and modalities, such as sound, imaging, proximity (sonar, radar), navigation, tactile,  etc. 
For the moment, we ignore the means by which we process and integrate the various signals, and simply refer to the given collection of signals as the ``raw signal.'' 
%Methods for learning features over multiple modalities using deep networks have been studied before 
%\citep{ngiam2011multimodal} and can be easily integrated in our scheme.
We propose the following scheme that integrates unsupervised feature learning, imitation learning, reward induction, safety module construction, and reinforcement learning as follows:
\begin{enumerate}
	\item
	{\bf Unsupervised learning.}  
	Utilizing known unsupervised learning techniques the agent learns informative low-dimensional representations $F_0(s)$ of the raw signal $s$.
	Typically, the representation is hierarchical, in the form of artificial neural network whose inputs are obtained from the raw signal and the output is the lower dimensional representation.

	{\bf External features.}
	In many tasks, such as autonomous driving, there are high level features of the raw signal that are known to be relevant to the task and accelerate the learning of the robot. 
	For example, road boundaries, which can be easily extracted using simple image processing algorithms (e.g., the Hough Transform \citep{illingworth1988survey}), are of utmost relevancy to the task. 
	Such engineered features that are based on domain expertise should definitely be used whenever they exist to expedite learning and/or improve the final performance.
	In addition, other auxiliary features that are not directly learned by our system, or engineered by us, and can obtained as black boxes from professional feature manufacturers can be considered.
	Any set of such \emph{external features} can be easily integrated into our scheme using known methods such as those described in \citep{ngiam2011multimodal}.
	\item
	{\bf Supervised imitation learning.}
	Using the learned low-dimensional (hierarchical) representation $F_0(s)$ (comprised of both self-learned features and possible external features) to initialize an imitation learning process, the agent learns to mimic an (human) instructor performing the desired task.
	This stage has two complementary goals.
	The first goal is to generate an initial agent policy $\pi_0$ capable of operating in the environment without too much risk (e.g., without damaging itself or the environment).
	The second goal is to improve the low-dimensional feature representation (learned in the previous unsupervised stage) and generate a revised representation $F_1(s)$, which is more relevant/informative to the task at hand.
	\item
	{\bf Supervised reward induction.}
	The agent \emph{learns} a reward function $R(s)$ (to be used later by the RL procedure) from instructor feedback generated while observing the agent operating in the environment using the initial IL policy $\pi_0$.
	The reward function learning process utilizes the learned representation, $F_1(s)$, to accelerate learning.
	\item
	{\bf Safety module construction.}
	The agent learns a \emph{safety function}, whose purpose is to classify the state space into two classes: safe and unsafe.
	During the RL training, if the safety function labels the current state as unsafe, the control is taken from the agent and is given to a pre-trained \emph{safe policy} until the agent is out of danger.
	We refer to the combination of a safety function plus a safe policy as a \emph{safety module}.
	The safety module learning process utilizes the learned representation, $F_1(s)$, to accelerate learning.
	\item
	{\bf Reinforcement learning.}
	Finally, in the RL stage, both the reward function $R(s)$ and the safety module are used to learn an improved agent policy $\pi^*$.	
	This learning in this stage can be based on the learned representation $F_1(s)$. 
	The representation $F_1(s)$ can remain ``frozen'' through the RL procedure or it can be updated to better reflect new scenes. 
	Also, the RL stage can remain active indefinitely. 
\end{enumerate}
We note that the first unsupervised feature learning stage is not mandatory, but can potentially accelerate the entire learning process and/or lead to enhanced overall performance.
When this stage is not conducted, we start with the supervised imitation learning stage from scratch, without $F_0(s)$, and when the supervised learning process ends, we extract from it both $F_1(s)$ and the initial robot policy $\pi_0$.
When the unsupervised feature learning stage is not conducted, the IL stage can take longer time and/or require more efforts from the human demonstrator.
We also note that stages three and five can be iteratively repeated several times to improve final performance.

% motivation for the paradigm
The proposed approach somewhat resembles the natural teaching-learning procedure used by humans to teach themselves and each other complex tasks.
For example, in the case of learning to drive, the student's unsupervised learning phase starts long before her formal driving lessons, and typically includes great many hours where the student observes driving scenes while sitting as a passive passenger in a car driven by her parents during her childhood.
In the second stage while observing the instructor performing a desired task, the student extracts relevant information required to successfully perform the task.
Afterwards, while the student is performing the task, the instructor provides real-time feedback and the student improves performance by both optimizing a policy as well as learning the feedback function itself. 
Then, the student continues to teach herself (without the instructor), using both the reward function previously induced by the instructor and future reward signals from the environment.
%.\footnote{To obtain a driver's license in California a 15-18 years old trainee is required to be instructed for 6 hours. An 18 years old trainee is not 
%	required for instruction at all (yet is required to pass a driving test). In Israel, 29 instruction 
%	hours are required for  
%In certain tasks, such as driving,  it was found that the feedback 
%stage should last sufficiently long (e.g., 29 driving hours with a teacher)
%Then, the student continues to teach herself (without the teacher), using both the reward function previously induced by the teacher and future reward signals from the environment. 

% close works to ours
There is vast literature concerning the independent use of each of the components in our scheme.
See, for example surveys on unsupervised feature learning \cite{coates2010analysis}, imitation learning \citep{argall2009survey} and reinforcement learning \citep{kober2013reinforcement}.
The two closest works to ours are \citep{taylor2011using} and \citep{daniel2014active}. 
In the first, the authors showed that a preceding demonstration learning stage can significantly expedite the reinforcement learning process and improve the final policy performance in a simulated robot soccer domain.
In the second, the authors proposed to learn a reward model in a supervised manner and used alternating steps of reward and reinforcement learning to continually improve the reward model and the agent's policy in a robotic arm grasping task.
To the best of our knowledge, there are no reports on previous attempts to leverage human instructor's skills to create robots using the above procedure (with or without the unsupervised learning stage, and with or without external features).
%   extract informative features during an imitation learning stage, and then utilize them for \emph{learning} a reward 
%model and conduct reinforecement learning.
 
% proof of concept
As a proof of concept, we focus on learning an autonomous highway steering task and report on an implementation of the proposed scheme, including the imitation learning, reward induction, safety module learning and reinforcement learning stages (without the unsupervised learning stage, and without external features, which we plan to add in a future version of the paper.)
To this end, we use one of the most popular computer racing games (Assetto Corsa)\footnote{At the time of writing, this game is considered to be one of the most realistic computer racing environments.}, and attempt to create a self-steering car in the sense that, given raw image pixels (of the car racing game screen), we wish to output correct steering control commands for the steering wheel. 
We use Convolutional Neural Networks (CNNs) as mapping functions from high-dimensional data to both control actions and instantaneous reward signals.
Four types of CNNs are trained: policy network, reward network, safety network, and Q-network, denoted by $P_{\theta}$, $R_{\theta}$, $S_{\theta}$ and $Q_{\theta}$, respectively.
The choice of CNNs for all three tasks is based on their proven ability to extract informative features from images in the context of classification and control tasks \citep{mnih2015human, krizhevsky2012imagenet}, thus obviating the exhausting task of manually defining features.
For example, in the work of \cite{mnih2015human} a CNN was successfully trained to predict desired control actions given high-dimensional pixel data in the Atari 2600 domain.

% no simulator
We emphasize that our entire system is implemented \emph{without} any access to the internal state of the game simulator (which is a purchased executable code).
This is in contrast to most previous published works on computerized autonomous driving, which were conducted in a simulation environment, allowing access to the internal states of the simulator (e.g., TORCS, \citealt{wymann2000torcs}).
These states contained valuable parameters such as the car's distance from the roadside or its angle with respect to the road. 
Thus, when an open simulator is available, such parameters can be extracted and utilized in the learning process, as can other reward information (e.g., in computer game simulators)  \citep{zhang2016query, loiacono2010learning, munoz2009controller, chen2015deepdriving}.  
Moreover, the lack of an open simulator in our setting means that all our learning procedures, including reinforcement learning, must be executed slowly in  wall-clock time (as in real driving), as opposed to the super-fast learning that can typically be achieved using a simulator.
However, overcoming this limitation in one game means that our method is in principle scalable to any computerized driving game.

% safe RL
We incorporate into the RL stage a novel method for reducing the number of catastrophic mistakes (i.e., car accidents) performed by the agent.
The goal of reducing critical mistakes made by an agent while interacting with an environment belongs to a subfield termed \emph{safe RL} \citep{garcia2015comprehensive}.
In safe RL, the learning agent seeks to find a policy that maximizes the long-term future reward received from the environment while respecting some safety constraints (e.g., damage avoidance).
Two of the main approaches for integrating the safety concept into the RL algorithms are: (i) transforming the optimization criteria to include a notion of risk (e.g., variance of return, worst-outcome); and (ii) modifying the agent's exploration process while utilizing prior knowledge of the task to avoid risky states.
Our proposed method is related to the second approach.
Using the labeled data of the human instructor, we learn a \emph{safety network}, whose purpose is to classify the state space into two classes: safe and unsafe.
During the RL training, if the safety network labels the current state as unsafe, the control is taken from the agent and given to a pre-trained \emph{safe policy} until the agent is out of danger.
By incorporating the safety module into the RL algorithm, we are able to significantly reduce the number of accidents (mostly in the initial stage of the RL).
Furthermore, the integration of the safety module into the RL allows us to improve the agent's exploration of the state space and speed up the learning, thus avoiding the pitfalls of random exploration policies such as $\epsilon$-greedy, where the agent wastes a lot of time on unimportant regions of the state space that the optimal policy would never have encountered.

% work four parts introduction
Our work is divided into four main parts.
First, the actions of a human demonstrator are recorded while she plays the game.
The game images are recorded as well.
A policy network is trained in a SL manner using this data.
Second, a reward network is learned.
It receives an image and outputs a number in the range $[-1,1]$ that indicates the instantaneous reward of being in that state.
We call this method \emph{reward induction} since the reward function is induced (in a SL manner) from a finite set of labeled examples obtained from a human instructor.
This method is related to a technique known as \emph{reward shaping}, where an additional reward signal is used to guide the learning agent \citep{ng1999policy}. 
Third, a safety module is constructed as previously explained, and is integrated into the RL stage.
Finally, the Double Deep Q-learning \citep{hasselt2010double, van2015deep} (DDQN) RL algorithm is used to train a Q-network.
The reward signal used in the RL procedure is constructed \emph{only} from the reward network's output.
The learned policy network's parameters from the imitation part are used to initialize the Q-network's parameters.
% performance evaluation 
Our performance evaluation procedure is based on the accumulated average reward.
Better driving means higher accumulated average reward and vice versa.

% optionally add an initial unsupervised stage
%An additional initial unsupervised feature learning stage, which can potentially accelerate the IL process, can be an integral part of the proposed scheme.
%In this stage the agent learns informative low-dimensional representation of the raw signal by utilizing known unsupervised learning techniques.
%Given unlabeled high-dimensional (and possibly multi-modal) data, which is received via the robot sensors while it performs the desired task, we aim at finding a function $U(s): s \to F_0(s)$, from the raw signal, $s$, to a low-dimensional representation, $F_0(s)$, of it.
%The learned feature representation, $F_0(s)$, is then used to accelerate the imitation stage (see below).
%One possible way to learn this feature representation is by utilizing autoencoding methods  \citep{bengio2009learning}.
%Denote by $U_{\theta}(s)$ the sub-network consisting of an autoencoder's first layers corresponding to the encoding transformation.
%When the autoencoder's training is done, this sub-network can be viewed as a mapping from the raw signal to its low-dimensional representation, $U_{\theta}(s): s \to F_0(s)$.
%When designing the architecture of the policy network, we set its first layers to be identical to the layers of $U_{\theta}(s)$. It is easy to utilize the learned low-dimensional representation obtained by 
%$U_{\theta}(s)$ and potentially accelerate the IL process by initializing the parameters of the first layers in the IL network with already learned parameters of $U_{\theta}(s)$.

\section{Imitation learning}
% introduction
Imitation learning, also known as behavioral cloning or learning from demonstrations, aims at finding a mapping $ f^{\star}: s \to a$ from a given world state, $s$, to a desired action $a$.
This mapping is typically termed  ``policy."
Robot learning using mimicry is an old idea, conceived decades ago \citep{hayes1994robot,argall2009survey}.
While being an excellent technique for achieving reasonable performance, this approach by itself is limited.
Clearly, the performance of an agent trained in this manner is upper bounded by the level of instruction.
Moreover, if the training sample isn't sufficiently diverse/representative, the agent will not be exposed to unexpected difficult states, and can suffer from very poor and unpredictable performance when such states are encountered in production.
Finally, labeled samples obtained from human demonstrations are prone to labeling noise.

Noting these limitations, we use imitation learning in our scheme only to achieve a basic performance level, which will allow the agent to perform the required task without damaging itself or the environment.
In our setting, world states are game images, and actions are keyboard keys pressed by a human demonstrator while playing the game.
The Assetto Corsa game offers the option of connecting a steering wheel controller; we have not utilized this option and note that the use of such a wheel controller should improve the driving performance of the learned agent.

% what we did
In this stage of the scheme we train a policy network that maps raw image pixels to steering commands of left/right.
The policy network, whose architecture is presented in Figure \ref{fig:policy_network_architecture}, is trained and evaluated.
A training sample of state-action pairs is gathered in the following procedure.

A training sample of state-action pairs, $D = \{(s_i,a_i)\}_i$, is recorded while a human expert plays the game.
A detailed  explanation of the recording process is given in Section \ref{sec: Eperimental section} (in the experiments described below $D$ contained approximately 70,000 samples, equivalent to two hours of human driving).
We then train a policy network using $D$, denoted by $P_{\theta}$.
The negative log-likelihood is used as a loss function for training $P_{\theta}$,
\begin{equation*}
NLL(\theta,{\cal{D}}) = - \frac{1}{|{\cal{D}}|}\sum_{i=0}^{|{\cal{D}}|} log P_{\theta} \left( a_{i}  | s_{i} \right) ,
%  NLL(\theta,{\cal{D}}) = \mathbb{E}_{s,a \sim \cal{D}} \left[ -\log P_{\theta} \left( a | s \right) \right] ,
\end{equation*}
where $P_{\theta}$ is a policy network, which receives the state $s$ and outputs a probability distribution on the optional actions, with parameters $\theta$.

A detailed explanation of the performance evaluation procedure is also given in Section \ref{sec: Eperimental section},
The parameters of $P_{\theta}$ were used to initialize the Q-network and the reward network.

\begin{figure} [h]
	\centering
	\begin{footnotesize}
		\begin{tabular}{ | c | }
			\hline
			Input - $6 \times 144 \times 192$   \\
			\hline
			\hline
			{\cellcolor{gray!10}}
			Conv1 - $ 6 \times 4 \times 4$  \\
			\hline
			{\cellcolor{gray!10}}
			Max Pooling $2 \times 2$   \\
			\hline
			\hline
			Conv2 - $ 8 \times 4 \times 4$   \\
			\hline
			Max Pooling $2 \times 2$   \\
			\hline
			\hline
			{\cellcolor{gray!10}}
			Conv3 - $ 16 \times 4 \times 4$   \\
			\hline
			{\cellcolor{gray!10}}
			Max Pooling $2 \times 2$   \\
			\hline
			\hline
			Conv4 - $ 16 \times 4 \times 4$   \\
			\hline
			Max Pooling $2 \times 2$   \\
			\hline
			\hline
			{\cellcolor{gray!10}}
			Fully Connected $100$   \\
			\hline
			\hline
			Fully Connected $3$   \\
			\hline
		\end{tabular}
	\end{footnotesize}
	\caption{
		The policy network's architecture.
		The nodes in the last layer correspond to the actions: 'No Action', 'Left' and 'Right'.
		Each convolutional layer is followed by a RELU nonlinearity.
		The numbers representing the convolutional layers and the Input layer are denoted for the number of channels $\times$ height $\times$ width.
		A softmax nonlinearity is applied on the last layer.
	}
	\label{fig:policy_network_architecture}
\end{figure}

\section{Deep reward induction}
\label{sec: reward netowrk}
% reward shaping introduction
The problem of designing suitable reward functions to guide an agent to successfully learn a desired task is known as \emph{reward shaping} \citep{laud2004theory}.
The idea is to define supplemental reward signals so as to make a RL problem easier to learn.
The handcrafting of a reward function can be a complicated task, requiring experience and a fair amount of specific domain knowledge.
Therefore, other methods for designing a reward function without the domain expertise requirement have been studied.
In \emph{Inverse RL} (IRL), a reward function is learned from expert demonstration \citep{abbeel2004apprenticeship}.
IRL algorithms rely on the fact that an expert demonstration implicitly encodes the reward function of the task at hand, and their goal is to recover a reward function which best explains the expert's behavior.

\citet{daniel2014active} proposed to learn a reward model in a supervised manner and use iterations between reward learning and reinforcement learning to continually improve the reward model and the agent's policy.
Their approach, which  is based on manual feature generation by experts, has been applied in a robotic arm grasping task. 
The reward function is not learned from the raw states visited by the learner, but from some assumed to be known, low-dimensional feature representation of them.
Constructing such low-dimensional representations usually requires some expert domain knowledge, making the learning of the reward function somewhat less advantageous.

% what we are doing
Our reward induction component \emph{learns} the reward function directly from the raw image pixels. 
Since we don't have access to the internal state of a simulator, defining a reward signal using the state parameters of the car (distances from the roadsides, angle with respect to the road, etc.) is impossible without explicit image processing.
In this work, we devise and utilize a deep reward network that is learned from a human driving instructor.
The reward network, which is implemented with convolutional layers, maps a game state into the instantaneous reward, $r \in [-1,1]$, corresponding to that state, $R_{\theta}: s \to r$.
The driving instructor provides binary labeling for each state such that the reward network is a mapping from raw image pixels to $\{$ ``good'', ``bad'' $\}$.
The binary labeling task, which is performed by the human instructor, continues until we are convinced that the reward model is sufficiently accurate (when evaluated on a test set.)

\begin{figure} [h]
	\centering
	\includegraphics[width=0.5\columnwidth]{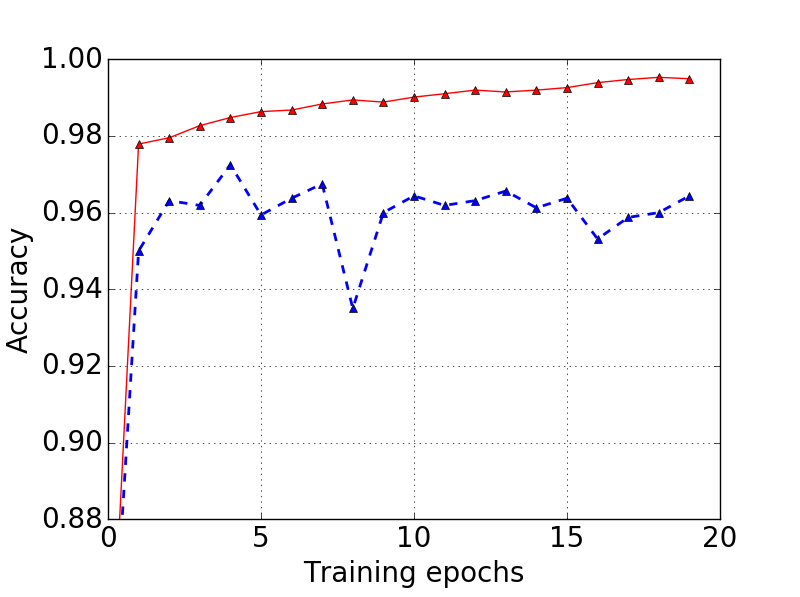}
	\caption{
		Training curve of $R_{\theta}^{lane}$.
		The red solid line represents training performance and the blue dashed line represents validation performance.
		$R_{\theta}^{lane}$ was trained on data generated from the ``Black Cat County" track.
	}
	\label{fig:SL_reward_lane_graphs}
\end{figure}

\begin{figure} [h]
	\begin{subfigure}{.49\columnwidth}
		\centering
		\includegraphics[width=0.9\columnwidth]{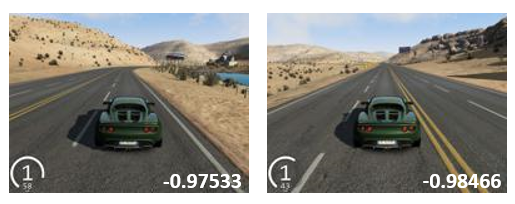}
		\caption{``Bad" states}
		\label{fig:lane_reward_bad}
	\end{subfigure}
	\begin{subfigure}{.49\columnwidth}
		\centering
		\includegraphics[width=0.9\columnwidth]{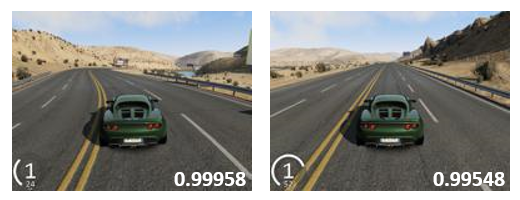}
		\caption{``Good" states}
		\label{fig:lane_reward_good}
	\end{subfigure}
	\caption{
		$R_{\theta}^{lane}$ outputs high reward to driving in the second lane from the right (b), and low reward to other lanes (a). 
		The reward is the number located in the lower-right corner of each frame.
		%		The samples are taken from the stage ``Black Cat County".
	}
	\label{fig:lane_reward}
\end{figure}

% additional information about the reward network 
During the imitation learning stage, the human instructor ignored the lane marks on the road (some of the tracks do not contain lane marks at all), but now we consider a more complex application whereby the agent must drive in a designated lane only (the second lane from the right on a 4-lane road).
To this end we train a reward network, denoted by $R_{\theta}^{lane}$, to give high reward only for states where the car was in that specific lane.
$R_{\theta}^{lane}$ has an identical architecture to that of the policy network, except for the final fully connected layer that now has one node instead of three. 
A hyperbolic tangent (tanh) activation function is applied on the last layer to receive output in the range $[-1,1]$.
The reward network is trained with data recorded from a human instructor in a supervised learning procedure.
The MSE loss objective is used to train the reward network,
\begin{equation*}
\mathbb{E}_{s,l \sim \cal{D}} \left[  \left( l-R_{\theta}\left(s \right) \right)^{2} \right] ,
\end{equation*}
where $l \in \{-1,1\}$ is the corresponding label given by the human instructor. 
$R_{\theta}^{lane}$ was trained with approximately 30,000 new samples, denoted by $D_{lane}$, corresponding to roughly one driving hour.
States in which the car was in the designated lane were labeled one, and all other states were labeled minus one.
A training curve of $R_{\theta}^{lane}$ is presented in Figure~\ref{fig:SL_reward_lane_graphs}, showing that a 97\% validation accuracy was achieved with early stopping.
An example of the output of $R_{\theta}^{lane}$ over different states is given in Figure \ref{fig:lane_reward}, showing very high reward to the designated lane.

\section{RL using DDQN}
\label{sec: RL using DDQN}
%  RL background
In this section we assume basic familiarity with reinforcement learning; see, e.g., \citep{sutton1998reinforcement}.
%TODO: Error! only one reward network! found it after the submisison.
In the final RL stage, we utilize the already trained reward networks and apply them within a standard RL method.  
Performance is measured in terms of the learned reward model, and the main goal of this stage is to achieve a significantly better performance level than the one achieved in the mimicry stage by enabling the agent to teach itself.
In other words, starting with a policy $\pi_0$, we would like to apply an RL algorithm to learn a policy $\pi^*$, which is optimal with respect to the expected (discounted) reward received from the reward network.

%  Q-network indroduction
We use a variant of the Q-learning algorithm \citep{watkins1992q}, which aims to find the \emph{optimal action-value function}, denoted by $Q^{\star}(s,a)$, and defined as the expected (discounted) reward after taking action $a$ in state $s$ and following the optimal policy $\pi^{\star}$ thereafter.
The true value of taking action $a$ in state $s$ under policy $\pi$ is
\begin{equation*}
Q^{\pi}(s,a) \! = \! \mathbb{E} \! \left[ r_{t} + \gamma \ \! r_{t+1} + \gamma^2 \ \! r_{t+2} + ...| s_t = s, \ \! a_t = a; \ \! \pi \right] \! ,
\end{equation*}
where $\gamma \in [0,1]$ is a fixed discount factor and $r$ is the guidance reward signal.
Given the optimal action-value function, which is defined to be $Q^{\star}(s,a) = \underset{\pi} {\operatorname{max}} \ \! Q^{\pi}(s,a)$, the optimal policy $\pi^{\star}$ can be simply derived by taking the action with the highest action-value function in each state. 
Dealing with a very large state space of images, we approximate the optimal action-value function using a \emph{deep Q-network} (DQN) with parameters $\theta$:
\begin{equation*}
Q_{\theta}(s,a) \approx Q^{\star}(s,a).
\end{equation*}
A deep Q-network is a neural network that for a given state outputs a vector of action values.

% algorithms being used
We used the DDQN algorithm with replay memory and target values calculated from parameters of the previous iteration, as in the work of \citet{mnih2015human}.
The loss function we used is therefore,
\begin{equation*}
\label{eq: Q-learning MSE}
L(\theta)  = \underset{s,a,r,s' \sim \cal{B}}{\mathbb{E}} \! \left[ \left( r+ \gamma \ \! Q_{\tilde{\theta}}(s',\underset{a'}{\operatorname{argmax}} \! \ Q_{\theta}(s',a')) - Q_{\theta}(s,a) \right)^{2} \right] \!\!,	
\end{equation*}
where $(s,a,r,s')$ are samples taken from the replay memory buffer $\cal{B}$, and $\tilde{\theta}$ are the Q-network's parameters from the previous iteration.
In our applications, the reward received after taking action $a$ in state $s$ is the instantaneous reward obtained from the reward network for state $s$ (see Section~\ref{sec: reward netowrk}). 
We set $\gamma$ to be $0.9$ and used the ADAM \citep{kingma2014adam} method for stochastic gradient optimization.

% what we did
We trained a Q-network, denoted by $Q_{\theta}$, that has the same architecture as $P_{\theta}$ except for the final softmax nonlinearity, which was removed.
The Q-network's parameters were initialized from those of the policy network except for the final layer's parameters, which were initialized from a uniform distribution $[-1 \times 10^{-3}, 1 \times 10^{3}]$.
We wished, on the one hand, to maintain the final convolutional layer of the policy network, which contains an informative state representation.
On the other hand, we also had to allow the Q-network to output values compatible with Q-learning temporal difference targets.
Therefore, an initial policy evaluation step was performed, where we let $P_{\theta}$ drive, and updated the parameters of only the last two fully-connected layers of $Q_{\theta}$ while fixing all of its other parameters.
This step can be viewed as learning a critic (and more precisely, only the last two layers of the critic's network) to estimate an action-value function while fixing the acting policy.
After this policy evaluation step, we started the RL algorithm using $Q_{\theta}$, allowing all of its parameters to be updated.
We refer to this type of initialization as \emph{IL initialization}.
Full details of the experiments are given in Section \ref{sec: Eperimental section}.

\subsection{Safety Network for Safe RL}
% what we did
We incorporate the notion of safety into the RL process using a \emph{safety module} that is composed of: (i) a safety network, denoted by $S_{\theta}$, that is able to identify risky states encountered by the agent; (ii) a safe driving policy, denoted by $P^{safety}_{\theta}$, that takes control of the agent's controller when it is in an unsafe state.
$S_{\theta}$ maps a game state into a real number in the range $[-1,1]$, corresponding to the amount of risk in that state, where the value $1$ is assigned to the safest possible state and the value $-1$ is assigned to the riskiest possible state.
States are classified as ``safe" and ``unsafe" according to the safety network's output; if it is above a predefined threshold, the state is labeled ``safe."
Otherwise it is labeled ``unsafe." 
When unsafe states are encountered, $P^{safety}_{\theta}$ takes control of the car until it is out of danger.
Specifically,
\begin{equation*}
\text{driving policy} =
\begin{cases} 
P^{safety}_{\theta}, & \text{if} \ S_{\theta} < threshold \\
Q_{\theta},			 & \text{if} \ S_{\theta} > threshold \ \! \text{.}\\
\end{cases}
\end{equation*}

\begin{figure} [h]
	\begin{subfigure}{.49\columnwidth}
		\centering
		\includegraphics[width=0.9\columnwidth]{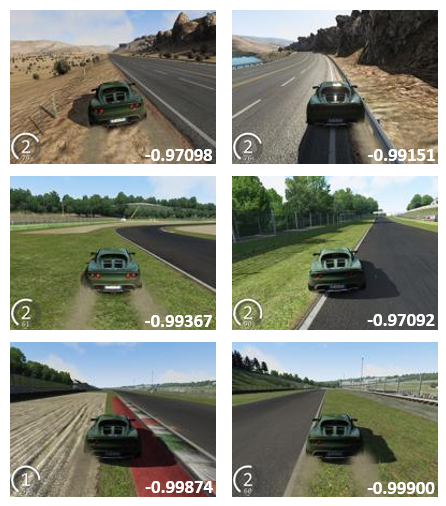}
		\caption{``Unsafe" states}
		\label{fig:default_reward_bad}
	\end{subfigure}
	\begin{subfigure}{.49\columnwidth}
		\centering
		\includegraphics[width=0.9\columnwidth]{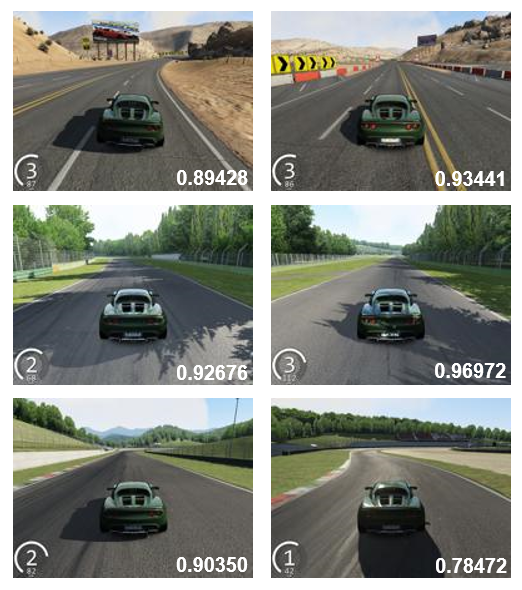}
		\caption{``Safe" states}
		\label{fig:default_reward_good}
	\end{subfigure}
	\caption{
		$S_{\theta}$ outputs a high score for ``safe" states according to the instructor (b), and low score for ``unsafe" states (a). 
		The output of the safety network is the number located in the lower-right corner of each frame.
	}
	\label{fig:default_reward}
\end{figure}

$S_{\theta}$ has the same architecture as $R_{\theta}^{lane}$, and is trained in the same way as $R_{\theta}^{lane}$, but with different dataset (including labeling).
Training data for $S_{\theta}$ was gathered in the following manner.
A human instructor drove the car while intentionally alternating between edge conditions (i.e., driving on, or even outside road boundaries) and safe driving; examples are provided in a supplementary video.
``Safe" states, where the car is on the road facing the correct direction, as demonstrated by the human instructor, are labeled one.
All other states are labeled minus one.
%\footnote{To ease the human instruction efforts, all of the one-labeled samples were reused from the imitation training data.}
The training of $S_{\theta}$ continues until a statistical test indicates that this network provides accurate results over test data.
The safe driving policy is the learned policy from the IL stage, $P_{\theta}$.
By using the proposed safety module, we were able to train our agent faster and with fewer accidents, as can be seen in Figures~\ref{fig:RL_results}~and~\ref{fig:random_initialization_with_safety}, and in Table~\ref{table:num_of_accidents}.
An example of the output of $S_{\theta}$ over different states is given in Figure \ref{fig:default_reward}.

% safety network figure

\begin{table}[t]
	\caption{
		Averaged number of accidents per epoch throughout the agent's RL training.
		The training is divided into four different episodes as follows: epochs 0-3, epochs 3-12, epochs 12-39 and epochs 39-120.
		An epoch is defined to be $15,000$ sample frames. % about 20 minutes
		The first row corresponds to a random initialization of the Q-network.
		In the second and third rows the Q-network is initialized with the IL policy parameters.
		In the third row the safety module is incorporated into the reinforcement learning.		
		The number of accidents in the last episode is averaged over 81 epochs, but in the final epochs both the $\textbf{IL}$ and $\textbf{IL + Safety}$ agents can complete tracks without accidents.
	}
	\label{table:num_of_accidents}
	\vskip 0.08in
	\begin{center}
		\begin{footnotesize}
			\begin{sc}
				%	\begin{tabular}{m{1.6cm} m{1.6em} m{1.6em} m{1.6em} m{1.6em}}
				\begin{tabular}{|l|cccc|}
					\hline
					%	&\Romannum{1} &\Romannum{2} &\Romannum{3} &\Romannum{4}\\
					& \textbf{0-3} & \textbf{3-12} & \textbf{12-39} & \textbf{39-120}\\
					\hline \hline
					\textbf{Random} 			&$116.6$		&$104.7$	&$68.5$		&$41$ 	\\
					\hline
					\textbf{IL}					&$67$			&$65.3$		&$68.1$		&$44.5$		\\
					\hline
					\textbf{IL + Safety}    	&$47$			&$46.1$		&$36.1$		&$23.3$ 		\\
					\hline
				\end{tabular}
			\end{sc}
		\end{footnotesize}
	\end{center}
	\vskip -0.1in
\end{table}

\section{Experiments and Results}
\label{sec: Eperimental section}
\begin{figure} [h]
	\begin{subfigure}{.49\columnwidth}
		\centering
		\includegraphics[width=1\columnwidth]{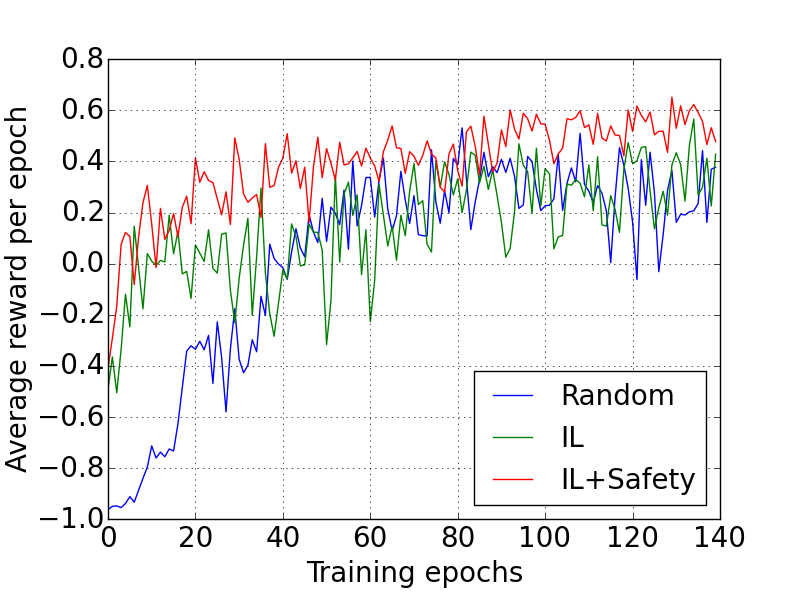}
		\vspace*{-0.57cm}
		\caption{
			\begin{small}
				Average reward
			\end{small}
		}
	\end{subfigure}
	\begin{subfigure}{.49\columnwidth}
		\centering
		\includegraphics[width=1\columnwidth]{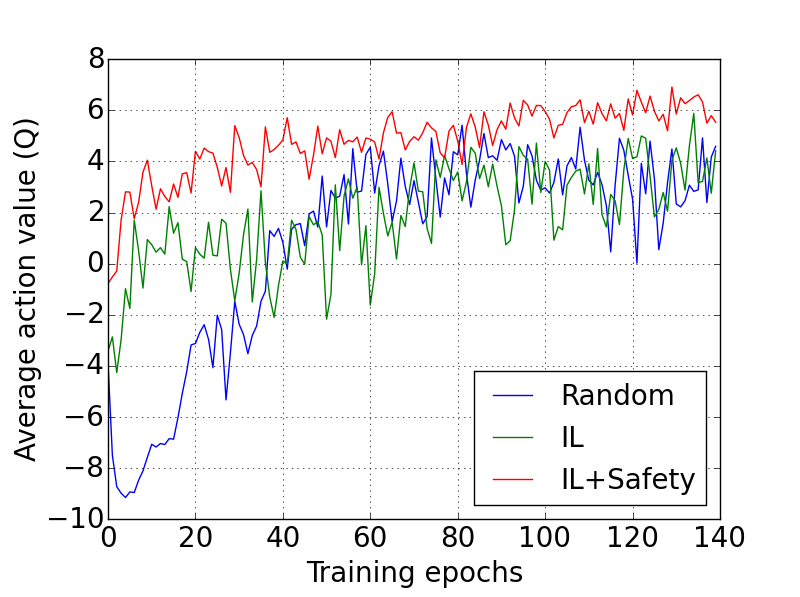}
		\vspace*{-0.57cm}
		\caption{
			\begin{small}
				Average action-value
			\end{small}	
		}
	\end{subfigure}
	\caption{
		Agent's training curves.
		An epoch is defined to be $15,000$ sample frames. % about 20 minutes
		The Q-network's parameters were initialized as follows:
		Blue: Random initialization.
		Green: IL initialization.
		Red: IL initialization + safety module.
		(a) Average reward per epoch.
		(b) Average action-value per epoch.
		The average reward achieved by the imitation policy, $P_{\theta}$, was $-0.45$.
	}
	\label{fig:RL_results}
\end{figure}

\begin{figure}[h]
	\centering
	\includegraphics[width=0.8\columnwidth]{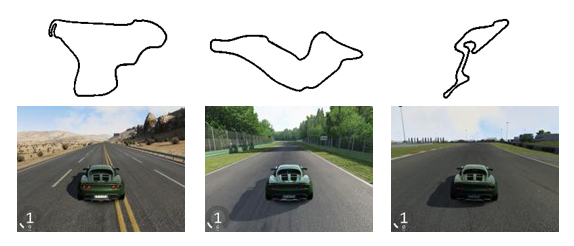}
	\caption{		
		The tracks we used with sample frames.
		All three tracks were used in the IL stage, and for learning the safety network.
		The tracks are (presented from left to right): ``Black Cat County", ``Imola" and ``N{\"u}rburgring gp".
		The ``Black Cat County'' track (left most) is the only one with lane marks, and it was used for the RL experiments.
	}
	\label{fig:sample_tracks}
\end{figure}

%  RL experiments and results
\textbf{RL experiments.} 
For the RL experiments we used $R_{\theta}^{lane}$ on the ``Black Cat County" track (see left most track in Figure~\ref{fig:sample_tracks}), which is the only track containing lane marks.
The Q-network's parameters were initialized in two different ways: (i) from the policy network's parameters, as explained in Section~\ref{sec: RL using DDQN}; and (ii) with random initialization for a baseline.
The agent operated with an $\epsilon$-greedy policy ($\epsilon = 0.05$).
The agent was trained for a total of 3.5 million frames (that is, around 4 days of game experience) with a replay memory of the $5,000$ most recent frames.
Figure~\ref{fig:RL_results} and Table~\ref{table:num_of_accidents} show that initializing the Q-network's parameters with those learned in the IL stage increased the learning rate at the beginning of the RL, and significantly reduced the number of critical driving mistakes made by the agent (compared to randomly initializing the Q-network).

\begin{figure}[h]
	\centering
	\includegraphics[width=0.5\columnwidth]{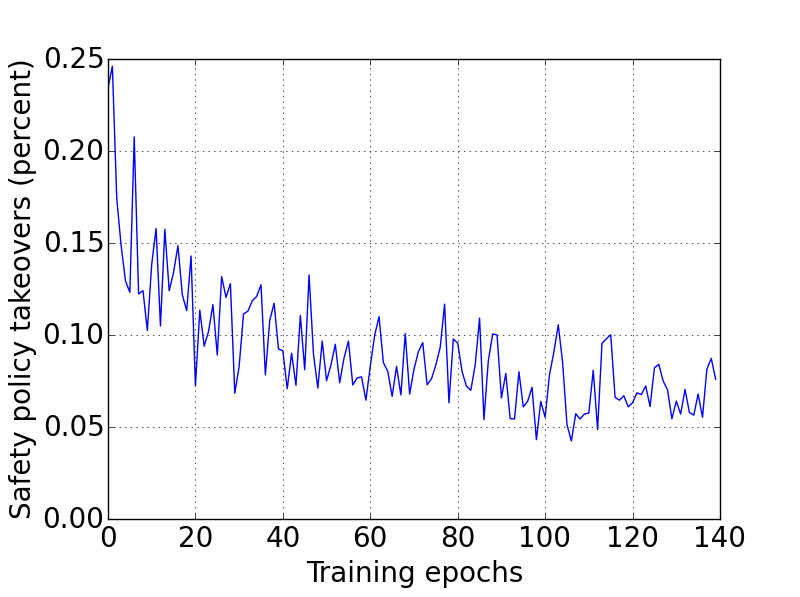}
	\caption{		
		Fraction of safety policy takeovers per decision in each epoch during the RL.
		An epoch is defined to be $15,000$ sample frames. % about 20 minutes
		Each point in the graphs is the relative number of sample frames per epoch in which the safety policy took control (i.e., the fraction of sample frames for which the output of the safety network was 'unsafe'.)
		The Q-network was initialized with the IL policy network parameters (IL initialization).
	}
	\label{fig:safety_policy_takeovers}
\end{figure}

\begin{figure}[h]
	\centering
	\includegraphics[width=0.5\columnwidth]{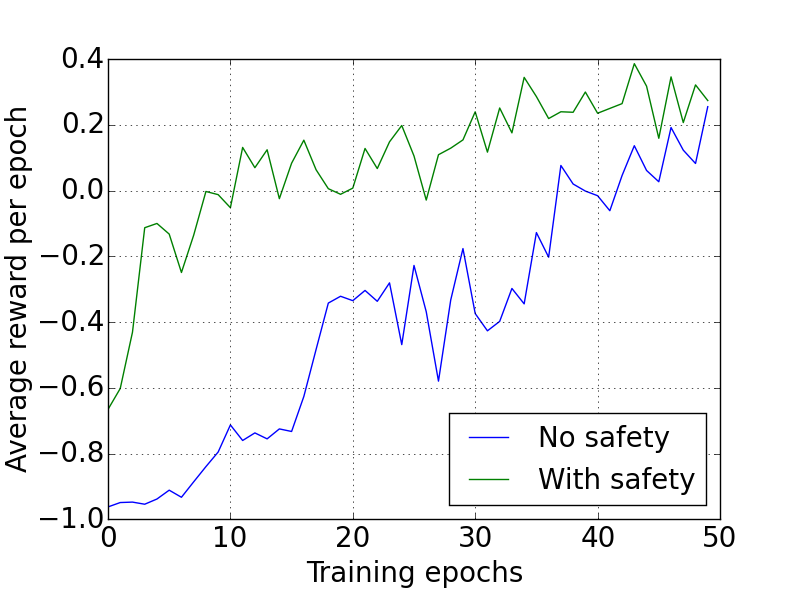}
	\caption{		
		Agent's training curves.
		An epoch is defined to be $15,000$ sample frames. % about 20 minutes
		Each point in the graph is the average reward per epoch.
		The Q-network was initialized with random parameters in both cases.
		Blue: no safety module.
		Green: with safety module.
	}
	\label{fig:random_initialization_with_safety}
\end{figure}

By incorporating the safety module into the RL process, we could further expedite the agent's learning rate and better circumvent critical driving mistakes. 
This is evident in Figure~\ref{fig:RL_results}, where the reward and action-value curves of $\textbf{IL + Safety}$ consistently dominates the corresponding $\textbf{IL}$ curves. 
Figure~\ref{fig:random_initialization_with_safety} further emphasizes the advantage of the safety module, now comparing randomly initialized RL with and without the safety module. 
Figure~\ref{fig:safety_policy_takeovers} presents the number of safe policy takeovers during the RL process; as time goes by the agent's driving skills improve and it is less exposed to dangerous states. 
The RL agent outperformed the human instructor's initial demonstration; namely, it received higher reward than the imitation policy $P_{\theta}$.

\begin{figure}[h]
	\centering
	\includegraphics[width=.5\columnwidth]{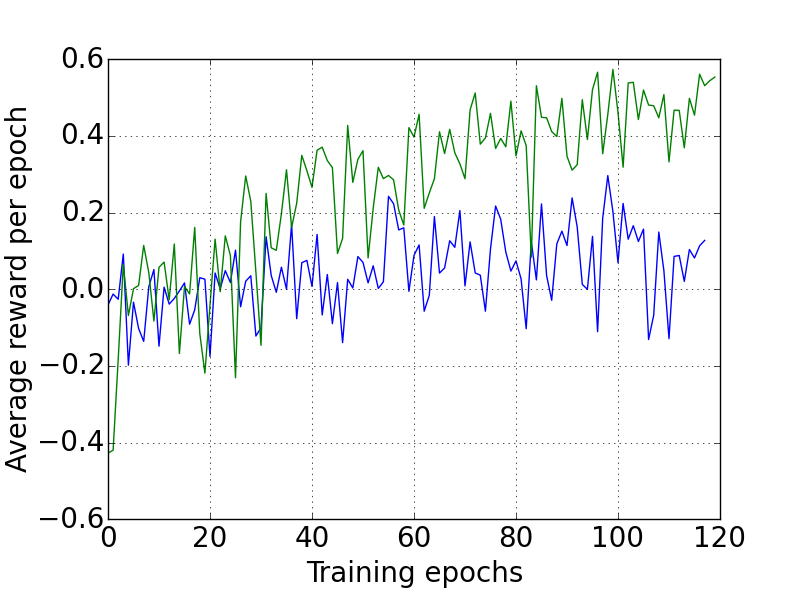}
	\caption{		
		Agent's RL training curves.
		An epoch is defined to be $30,000$ sample frames. % about 40 minutes
		Each point in the graphs is the average reward achieved per epoch from the reward network $R_{\theta}^{lane}$.
		Green: reward signal is received from $R_{\theta}^{lane}$. 
		Blue: reward signal is received from $R_{\theta}^{\ \! 0.2 \ \! lane}$.	
	}
	\label{fig:BlackCat_reward_partial_data}
\end{figure}

A high quality reward signal is critical to the agent's success during the RL stage.
In the proposed scheme the reward signal is learned from human instructions.
Therefore, the learned reward network must be able to generalize the human instructions for unseen states.
To investigate the effect of the reward signal's quality on the agent's overall performance during the RL stage, we conducted the following experiment.
An additional reward network, denoted by $R_{\theta}^{\ \! 0.2 \ \! lane}$, was trained using only $20\%$ of the dataset $D_{lane}$, achieving $66\%$ validation accuracy.
We then trained an RL agent with a reward signal given from $R_{\theta}^{\ \! 0.2 \ \! lane}$ and compared its performance to the RL agent trained using $R_{\theta}^{lane}$.
The results, presented in Figure~\ref{fig:BlackCat_reward_partial_data}, emphasize the necessity of a reward function that is able to correctly generalize the human instructor's knowledge to unseen states.

%For Task 2, two additional reward networks were trained using smaller datasets than $D_{lane}$. %TODO: improve sentece!!
%The first, $R_{\theta}^{0.2 lane}$, was trained using $20\%$ of the dataset $D_{lane}$, achieving $66\%$ validation accuracy.
%The second, $R_{\theta}^{0.6 lane}$, was trained using $60\%$ of the dataset $D_{lane}$, achieving $84\%$ validation accuracy.
%The experiment's results, which are presented in Figure~\commentBar{add figure}, indicates on an important property a ``good" reward function should possess - smoothness.
%% TODO: improve these two sentences
%As expected, the agent that received a reward signal from $R_{\theta}^{0.2 lane}$ performed worse than the one that received a reward signal from $R_{\theta}^{lane}$.
%Whereas, surprisingly, the agent that received a reward signal from $R_{\theta}^{0.6 lane}$ performed better than the one that received a reward signal from $R_{\theta}^{lane}$.
%Our explanation for this phenomena is given in Figure~\commentBar{add figure}.
%%TODO: improve sentence
%A measure for the different reward networks smoothness is given in Figure~\commentBar{add figure} were a histogram of the different reward networks outputs during the agent's RL training is presented.

% Working without a simulator
\textbf{Technical implications of not having access to the simulator's internal state.}
While performing the RL algorithm, the following operations are executed online: image capturing, image pre-processing, reward prediction by the reward network, action prediction by the Q-network, the Q-network parameter updates, etc. 
On average, the serial execution of these operations takes $70$ ms.
Therefore, the agent must always act under $70$ ms latency (from the time the image is captured to the time the corresponding steering key is pressed). 
During this gap, the previous key continues to be pressed until a new one is received.
This added difficulty can be avoided when using a simulator.
But most importantly, our inability to apply the RL stage at a super-fast simulation speed limits the number of training epochs we can afford to conduct.

% preprocessing
\textbf{Image preprocessing.}
The images were resized from $1024 \times 768$ to $192 \times 144$ and were not converted to gray scale.
The pixel values were scaled to be in the range $[0,1]$. 
Pixels corresponding to the speed indicator were set to zero in order to guarantee that the agent doesn't use the speed information during the learning process.
In all our experiments, each input instance consisted of two sequential images with a $0.5$-second gap between them. 

% work environment
\textbf{Work environment.}
We used the Theano-based Lasagne library for implementing the neural networks.
Two GeForce GTX TITAN X GPUs were utilized during the RL experiments; one was used to run the racing game and the other to train the networks and predict the rewards and actions. 

% data collecting
\textbf{Data generation and network training technicalities.} 
The human demonstrator played the game using a ``racing" strategy: driving as fast as possible while ignoring the lane marks and trying to avoid accidents. 
Training data for the policy and reward networks was collected from the tracks: 'Black Cat County', 'Imola' and 'N{\"u}rburgring gp'.
Sample images of the tracks we used are presented in Figure~\ref{fig:sample_tracks}.
Using a Python environment, screen images were recorded every $0.1$ seconds while the human demonstrator played the game.
Keyboard keys pressed by the demonstrator were also recorded.
The same sampling rate was used when gathering data for the reward induction stage.
$80$\% of the recorded samples were used as training data and the remaining $20$\% as validation data.
We used the ADAM stochastic optimization method with dropout for regularization \citep{srivastava2014dropout}.
The network parameters that achieved best validation accuracy were chosen.

% performamce measure
\textbf{Performance evaluation.}
Without a natural performance evaluation measure for driving skills and without access to the internal state of the game, our performance evaluation procedure was based on the accumulated average reward achieved by the agent.
Better driving means higher (discounted) accumulated average reward and vice versa.
Assuming that our learned reward model reliably reflects the concept of ``good driving'' (with or without lane constraints), the accumulated  average reward is a sound performance measure for the required task.
% A video illustrates the car's driving can be viewed in \comment{add video.}  %  add video 

% Other technicalities.
\textbf{Other technicalities.}
Game restarts were performed when the agent had zero speed (e.g., stuck against some obstacle) or when it drove in the wrong direction (both of these situations were detected through simple image processing).

%We detect a car accident if the difference between two sequential speed frames exceeds a predefined threshold of $10$ kmh.
All of the experiments were conducted without any other vehicles on the road.
We used the 'Lotus Elise SC' car model in all of our experiments.
Focusing only on the steering control problem, we eliminated the accelerator/brakes control variability by using a ``cruise control" behavior whereby the car's speed was set to 50 kmh. 
This was achieved by extracting the car's speed from the game image and applying a simple hand-crafted controller.

\section{Concluding remarks}
% what we did
We presented a generic learning framework consisting of five elements that allows for end-to-end training of a computerized agent (without using any image processing or other hand-crafted features) to perform a complex task, using only raw visual signals.
We expect the proposed framework to be useful in various application domains, and have demonstrated its strength on an autonomous highway steering problem. 
Our solution relies on recent deep representational methods (CNNs) to successfully implement four elements of the proposed framework (without the first unsupervised learning stage).
We first train a CNN agent to imitate a human demonstrator, with the goal of achieving a basic initial driving policy.
An instantaneous reward network is then induced from human feedback.
A novel safety module, composed of a safety network and a safe policy, is integrated into the RL framework. 
The agent then uses the learned reward network's output as a guidance signal in the RL procedure. 
We have successfully demonstrated the advantage of all four elements, and we note that the proposed implementation is scalable to any computerized (black-box) driving game where the steering is controlled via the keyboard.

The main novelty of this work is that it enables to utilize the supervision and instruction abilities of a human to train a robotic agent to perform complex tasks.
The instructor should be able to perform the task himself only at a very basic level (and demonstrate it to the agent), and should be able to provide qualifying feedback to the agent if actions taken by the agent are advancing the task or not. 
When safety module construction is applied, the instructor should demonstrate safe/unsafe actions. 
The proposed method enables in principle the creation of an agent capable of performing tasks that the instructor cannot perform herself.

A typical real-world RL application often relies on a accurate simulator of the environment, which enables super-fast applications of RL algorithms by accelerating the real-world clock.
In our setting it is impossible to accelerate the real-world clock (as in many other real-world tasks for which constructing an accurate simulator is extremely difficult).
This limitation is hard to overcome and requires appropriate and extremely effective RL, as we aim to achieve here.
In contrast, in the work of \citet{mnih2015human}, where RL is used to predict desired control actions in the Atari 2600 domain, it is possible to train the RL agent (using the open source ALE simulator) 100 times faster than in our setting. 

An additional strength of our scheme lies in leveraging the weak supervision abilities of a (human) instructor, who, while unable to perform well herself at the required task, can provide coherent and learnable instantaneous reward signals to the computerized trainee. 
This leveraging effect clearly occurred in our self-steering example, where single-lane driving demonstrations (by the instructor) were not included in the imitation stage (and moreover, most of the training tracks do not even include lane marks). 
Nonetheless, the agent quite easily learned to drive in a single designated lane after a one-hour instruction session followed by self reinforcement learning (without the instructor).

The proposed safety module proved itself as an extremely effective technique to expedite the RL in our setting.
Our supervised approach to construct this module was dependent on the particular task we considered.
In general, it would be very interesting to develop effective approaches to constructing such modules for arbitrary vision-based tasks, perhaps using semi-supervised or even unsupervised techniques.

We believe that the proposed approach can be used to train robotic agents, in the absence of realistic simulator in real-world environments. 
To this end, both effective acquisition of instantaneous rewards from an instructor and accurate modeling of the reward function are required for successful application of the proposed scheme.
While in our driving example the reward models were easily constructed, creating these models for highly complex tasks is expected to be challenging in terms of both model capacity and the development of effective methodologies for interaction with the instructor.
To handle more complex tasks (e.g., cooking, dish washing) the policies must operate a number of controllers (or more complicated controllers with many degrees of freedom). 
An interesting question in this regard is how to construct appropriate network architectures and training methodologies to jointly handle many related controllers.
In our context it would be interesting to extend the proposed solution to handle the accelerator/brakes controllers. 
We anticipate that harnessing the supervision abilities of a (human) instructor, for the purpose of learning an effective reward model, will become a critical building block in creating robots capable of adjusting themselves to human needs.

%@techreport{hayes1994robot,
%	title={A Robot Controller Using Learning by Imitation},
%	author={G.~M. Hayes and J.~Demiris},
%	year={1994},
%	publisher={University of Edinburgh, Department of Artificial Intelligence}
%}

\bibliographystyle{iclr2017_conference}
\bibliography{references}

\begin{thebibliography}{25}
\providecommand{\natexlab}[1]{#1}
\providecommand{\url}[1]{\texttt{#1}}
\expandafter\ifx\csname urlstyle\endcsname\relax
  \providecommand{\doi}[1]{doi: #1}\else
  \providecommand{\doi}{doi: \begingroup \urlstyle{rm}\Url}\fi

\bibitem[Abbeel \& Ng(2004)Abbeel and Ng]{abbeel2004apprenticeship}
P.~Abbeel and A.~Y. Ng.
\newblock {Apprenticeship Learning via Inverse Reinforcement Learning}.
\newblock In \emph{Proceedings of the Twenty-First International Conference on
  Machine Learning}, 2004.

\bibitem[Argall et~al.(2009)Argall, Chernova, Veloso, and
  Browning]{argall2009survey}
B.~D. Argall, S.~Chernova, M.~Veloso, and B.~Browning.
\newblock {A survey of robot learning from demonstration}.
\newblock \emph{Robotics and Autonomous Systems}, 57\penalty0 (5):\penalty0
  469--483, 2009.

\bibitem[Chen et~al.(2015)Chen, Seff, Kornhauser, and
  Xiao]{chen2015deepdriving}
C.~Chen, A.~Seff, A.~Kornhauser, and J.~Xiao.
\newblock {Deepdriving: Learning Affordance for Direct Perception in Autonomous
  Driving}.
\newblock In \emph{Proceedings of the IEEE International Conference on Computer
  Vision}, pp.\  2722--2730, 2015.

\bibitem[Coates et~al.(2010)Coates, Lee, and Ng]{coates2010analysis}
Adam Coates, Honglak Lee, and Andrew~Y Ng.
\newblock {An analysis of single-layer networks in unsupervised feature
  learning}.
\newblock \emph{Ann Arbor}, 1001\penalty0 (48109):\penalty0 2, 2010.

\bibitem[Daniel et~al.(2014)Daniel, Viering, Metz, Kroemer, and
  Peters]{daniel2014active}
C.~Daniel, M.~Viering, J.~Metz, O.~Kroemer, and J.~Peters.
\newblock {Active Reward Learning}.
\newblock In \emph{Proceedings of Robotics Science \& Systems}, 2014.

\bibitem[Garc{\i}a \& Fern{\'a}ndez(2015)Garc{\i}a and
  Fern{\'a}ndez]{garcia2015comprehensive}
Javier Garc{\i}a and Fernando Fern{\'a}ndez.
\newblock {A comprehensive survey on safe reinforcement learning}.
\newblock \emph{Journal of Machine Learning Research}, 16\penalty0
  (1):\penalty0 1437--1480, 2015.

\bibitem[Hasselt(2010)]{hasselt2010double}
H.~Van Hasselt.
\newblock {Double Q-learning}.
\newblock In \emph{Advances in Neural Information Processing Systems}, pp.\
  2613--2621, 2010.

\bibitem[Hasselt et~al.(2016)Hasselt, Guez, and Silver]{van2015deep}
H.~Van Hasselt, A.~Guez, and D.~Silver.
\newblock {Deep Reinforcement Learning with Double Q-Learning}.
\newblock In \emph{Proceedings of the Thirtieth AAAI Conference on Artificial
  Intelligence}, 2016.

\bibitem[Hayes \& Demiris(1994)Hayes and Demiris]{hayes1994robot}
G.~M. Hayes and J.~Demiris.
\newblock {A Robot Controller Using Learning by Imitation}.
\newblock Technical report, University of Edinburgh, Department of Artificial
  Intelligence, 1994.

\bibitem[Illingworth \& Kittler(1988)Illingworth and
  Kittler]{illingworth1988survey}
John Illingworth and Josef Kittler.
\newblock {A survey of the Hough transform}.
\newblock \emph{Computer vision, graphics, and image processing}, 44\penalty0
  (1):\penalty0 87--116, 1988.

\bibitem[Kingma \& Ba(2014)Kingma and Ba]{kingma2014adam}
D.~Kingma and J.~Ba.
\newblock {Adam: A Method for Stochastic Optimization}.
\newblock \emph{arXiv preprint arXiv:1412.6980}, 2014.

\bibitem[Kober et~al.(2013)Kober, Bagnell, and Peters]{kober2013reinforcement}
J.~Kober, J.~A. Bagnell, and J.~Peters.
\newblock {Reinforcement Learning in Robotics: A survey}.
\newblock \emph{The International Journal of Robotics Research}, 2013.

\bibitem[Krizhevsky et~al.(2012)Krizhevsky, Sutskever, and
  Hinton]{krizhevsky2012imagenet}
A.~Krizhevsky, I.~Sutskever, and G.~E. Hinton.
\newblock {Imagenet Classification with Deep Convolutional Neural Networks}.
\newblock In \emph{Advances in Neural Information Processing Systems}, pp.\
  1097--1105, 2012.

\bibitem[Laud(2004)]{laud2004theory}
A.~D. Laud.
\newblock \emph{{Theory and Application of Reward Shaping in Reinforcement
  Learning}}.
\newblock PhD thesis, University of Illinois at Urbana-Champaign, 2004.

\bibitem[Loiacono et~al.(2010)Loiacono, Prete, Lanzi, and
  Cardamone]{loiacono2010learning}
D.~Loiacono, A.~Prete, P.~L. Lanzi, and L.~Cardamone.
\newblock {Learning to Overtake in TORCS Using Simple Reinforcement Learning}.
\newblock In \emph{IEEE Congress on Evolutionary Computation}, pp.\  1--8,
  2010.

\bibitem[Mnih et~al.(2015)Mnih, Kavukcuoglu, Silver, Rusu, Veness, Bellemare,
  Graves, Riedmiller, Fidjeland, Ostrovski, et~al.]{mnih2015human}
V.~Mnih, K.~Kavukcuoglu, D.~Silver, A.~A. Rusu, J.~Veness, M.~G. Bellemare,
  A.~Graves, A.~Riedmiller, A.~K. Fidjeland, G.~Ostrovski, et~al.
\newblock {Human-level control through deep reinforcement learning}.
\newblock \emph{Nature}, pp.\  529--533, 2015.

\bibitem[Munoz et~al.(2009)Munoz, Gutierrez, and Sanchis]{munoz2009controller}
J.~Munoz, G.~Gutierrez, and A.~Sanchis.
\newblock {Controller for TORCS created by imitation}.
\newblock In \emph{IEEE Symposium on Computational Intelligence and Games},
  pp.\  271--278, 2009.

\bibitem[Ng et~al.(1999)Ng, Harada, and Russell]{ng1999policy}
A.~Y. Ng, D.~Harada, and S.~J. Russell.
\newblock {Policy invariance under reward transformations: Theory and
  application to reward shaping}.
\newblock In \emph{International Conference on Machine Learning}, volume~99,
  pp.\  278--287, 1999.

\bibitem[Ngiam et~al.(2011)Ngiam, Khosla, Kim, Nam, Lee, and
  Ng]{ngiam2011multimodal}
Jiquan Ngiam, Aditya Khosla, Mingyu Kim, Juhan Nam, Honglak Lee, and Andrew~Y
  Ng.
\newblock {Multimodal deep learning}.
\newblock In \emph{Proceedings of the 28th international conference on machine
  learning (ICML-11)}, pp.\  689--696, 2011.

\bibitem[Srivastava et~al.(2014)Srivastava, Hinton, Krizhevsky, Sutskever, and
  Salakhutdinov]{srivastava2014dropout}
N.~Srivastava, G.~Hinton, A.~Krizhevsky, I.~Sutskever, and R.~Salakhutdinov.
\newblock {Dropout: a simple way to prevent neural networks from overfitting}.
\newblock \emph{Journal of Machine Learning Research}, 15\penalty0
  (1):\penalty0 1929--1958, 2014.

\bibitem[Sutton \& Barto(1998)Sutton and Barto]{sutton1998reinforcement}
R.~S. Sutton and A.~G. Barto.
\newblock \emph{{Reinforcement Learning: An Introduction}}.
\newblock MIT press Cambridge, 1998.

\bibitem[Taylor et~al.(2011)Taylor, Suay, and Chernova]{taylor2011using}
M.~E. Taylor, H.~B. Suay, and S.~Chernova.
\newblock {Using Human Demonstrations to Improve Reinforcement Learning}.
\newblock In \emph{AAAI Spring Symposium: Help Me Help You: Bridging the Gaps
  in Human-Agent Collaboration}, 2011.

\bibitem[Watkins \& Dayan(1992)Watkins and Dayan]{watkins1992q}
C.~J. C.~H. Watkins and P.~Dayan.
\newblock {Q-Learning}.
\newblock \emph{Machine Learning}, pp.\  279--292, 1992.

\bibitem[Wymann et~al.(2000)Wymann, Espi{\'e}, Guionneau, Dimitrakakis, Coulom,
  and Sumner]{wymann2000torcs}
B.~Wymann, E.~Espi{\'e}, C.~Guionneau, C.~Dimitrakakis, R.~Coulom, and
  A.~Sumner.
\newblock {TORCS, the open racing car simulator}.
\newblock \emph{Software available at http://torcs. sourceforge. net}, 2000.

\bibitem[Zhang \& Cho(2016)Zhang and Cho]{zhang2016query}
J.~Zhang and K.~Cho.
\newblock {Query-Efficient Imitation Learning for End-to-End Autonomous
  Driving}.
\newblock \emph{arXiv preprint:1605.06450}, 2016.

\end{thebibliography}

%\section{Appendix}

\end{document}